\documentclass[sigconf]{acmart}
\pdfoutput=1
\AtBeginDocument{%
  }

\setcopyright{acmlicensed}
\copyrightyear{2025}
\acmYear{2025}
\acmDOI{XXXXXXX.XXXXXXX}
\acmConference[MM2025]{Make sure to enter the correct
  conference title from your rights confirmation email}{October 27-0ctober 31,2025}{Dublin, lreland}
\acmISBN{978-1-4503-XXXX-X/2018/06}

\usepackage{graphicx}
\begin{document}

\title{EHPE: A Segmented Architecture for Enhanced Hand Pose Estimation}


\author{Bolun Zheng}
\affiliation{%
  \institution{Hangzhou Dianzi University}
  \city{Hangzhou}
  \state{Zhejiang}
  \country{China}}

\author{Xinjie Liu}
\affiliation{%
 \institution{Hangzhou Dianzi University}
  \city{Hangzhou}
  \state{Zhejiang}
  \country{China}
}
\email{232060254@hdu.edu.cn}

\author{Qianyu Zhang}
\affiliation{%
 \institution{Hangzhou Dianzi University}
  \city{Hangzhou}
  \state{Zhejiang}
  \country{China}}

\author{Canjin Wang*}
\affiliation{%
 \institution{Xinhua Zhiyun Technology Co., Ltd.}
  \city{Hangzhou}
  \state{Zhejiang}
  \country{China}}

\author{Fangni Chen}
\affiliation{%
  \institution{Xinhua Zhiyun Technology Co., Ltd.}
  \city{Hangzhou}
  \state{Zhejiang}
  \country{China}}

\author{Mingen Xu*}
\affiliation{%
  \institution{Hangzhou Dianzi University}
  \city{Hangzhou}
  \state{Zhejiang}
  \country{China}
}

\renewcommand{\shortauthors}{Zheng et al.}

\begin{abstract}
3D hand pose estimation has garnered great attention in recent years due to its critical applications in human-computer interaction, virtual reality, and related fields.
The accurate estimation of hand joints is essential for high-quality hand pose estimation.
However, existing methods neglect the importance of Distal Phalanx Tip (TIP) and Wrist in predicting hand joints overall and often fail to account for the phenomenon of error accumulation for distal joints in gesture estimation, which can cause certain joints to incur larger errors, resulting in misalignments and artifacts in the pose estimation and degrading the overall reconstruction quality.
To address this challenge, we propose a novel segmented architecture for enhanced hand pose estimation (EHPE).
We perform local extraction of TIP and wrist, thus alleviating the effect of error accumulation on TIP prediction and further reduce the predictive errors for all joints on this basis.
EHPE consists of two key stages: In the TIP and Wrist Joints Extraction stage (TW-stage), the positions of the TIP and wrist joints are estimated to provide an initial accurate joint configuration; In the Prior Guided Joints Estimation stage (PG-stage), a dual-branch interaction network is employed to refine the positions of the remaining joints. 
Extensive experiments on two widely used benchmarks demonstrate that EHPE achieves state-of-the-arts performance.

\end{abstract}



\received{23 April 2025}
\received[revised]{19 June 2025}
\received[accepted]{4 July 2025}

\maketitle

\setlength{\intextsep}{0pt}
\setlength{\textfloatsep}{0pt}
\begin{figure}[!htbp]
  \includegraphics[width=\columnwidth]{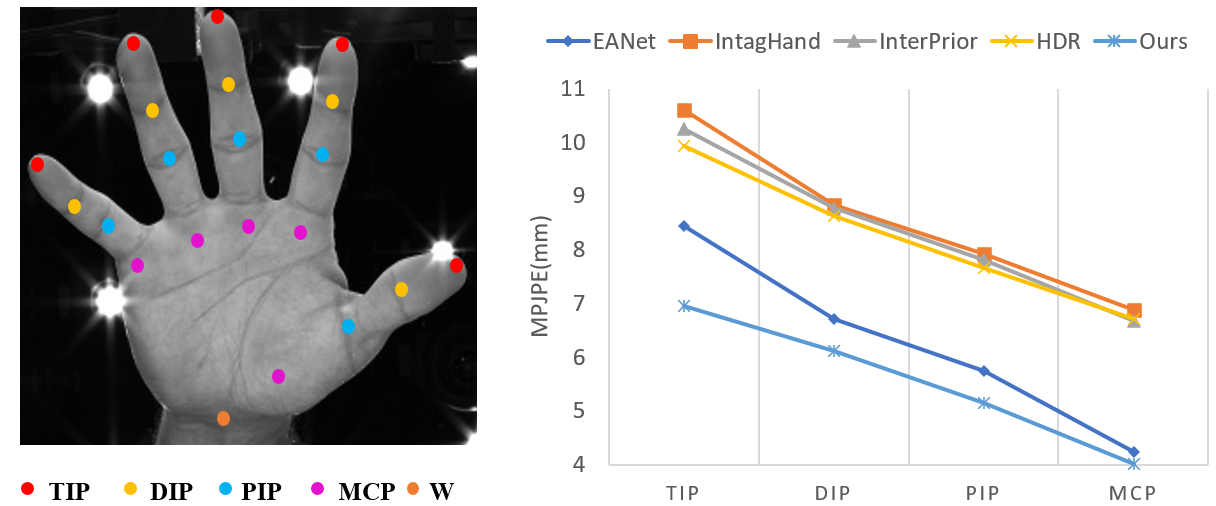}
  \caption{Compared to other joint errors in existing methods, the error of TIP is significantly higher than that in the remaining joints.}
  \Description{Enjoying the baseball game from the third-base
  seats. Ichiro Suzuki preparing to bat.}
  \label{icml-prem_1}
\end{figure}

\section{Introduction}
 Hand pose estimation is an essential problem in computer vision with broad applications in human-computer interaction, virtual reality, and sign language recognition \cite{Ge_Liang_Yuan_Thalmann_2017, wu2005analyzing, hampali2022keypoint, Liang_Yuan_Lee_Ge_Thalmann_2019,he20213d,zuo2023reconstructing}. 
 Despite significant advancements in this field, hand pose estimation remains challenging, primarily due to the diversity of hand poses and complex self-occlusions.
 Traditionally, hand pose estimation can be divided into two key tasks: 3D hand pose estimation and mesh regression. 
 Accurate 3D hand pose estimation is crucial, as it is a foundation for high-quality hand mesh reconstruction.

With the rapid development of deep learning techniques, deep learning-based methods \cite{deng2022recurrent,huang2020awr,cai2018weakly,baek2019pushing,wang2024rethinking,meng20223d,jiang2023a2j} such as heatmap-based methods, point cloud-based methods, etc., have become mainstream and show promising abilities to handle complex poses, occlusions, backgrounds and lighting environments.
These methods always utilize pre-trained backbone (\textit{i.e.} ResNet \cite{he2016deep}, ViT \cite{vasu2023fastvit}) to extract visual features and use carefully designed detectors to estimate hand poses \cite{zhang2024handformer2t, zhou2024simple}.
Recently, researchers have noted that prior knowledge of hand structure can be involved using the Graphic Convolution Network (GCN) \cite{kipf2016semi}, to improve the accuracy of hand pose estimation \cite{yao2024decoupling,ren2023decoupled}.
 

Although their promising performances, they still suffered from instability, particularly when dealing with complex curved joints.
We conducted a quantitative investigation on several recently proposed methods, and visualized the results in Figure \ref{icml-prem_1}.
In the investigation, we first categorize hand joints into five distinct classes: Distal Phalanx Tip (TIP), Distal Interphalangeal Joint (DIP), Proximal Interphalangeal Joint (PIP), Metacarpophalangeal Joint (MCP) and Wrist(W), as suggested by \cite{spurr2020weakly}.
Then we measure the average errors of every category produced by different methods.
There are two interesting observations: 
\begin{itemize}
    \item All methods exhibit the largest error for TIP, with the magnitude following the order: TIP> DIP> PIP> MCP> W.
    \item With an error of 100\% for TIP, all methods show similar relative errors for other four categories, approximately 80\% for DIP, 72\% for PIP, 59\% for MCP, and 40\% for W.
\end{itemize}
Our analysis suggests that this phenomenon can be attributed to the following reasons:
First, the wrist has only one joint with distinct feature, which makes it relatively easy to estimate. 
In contrast, TIP joints are numerous and each of them get closed visual characteristics to others. 
Besides, they are of higher susceptibility to occlusion, which further results in greater challenges for accurate estimation.
Second, in the natural state, human hands can only freely control the position of TIP.
Once the positions of TIP relative to the wrist are determined, the relative positions of other joints can almost be locked.
Thus, the prior knowledge of hand structure can be further details as: the positions of DIP, PIP and MCP are determined by the positions of TIP and wrist.
Therefore, TIP estimation errors can be regarded as the source of overall hand pose estimation errors. 

Following above analysis, we propose a segmented architecture for Enhanced Hand Pose Estimation (EHPE).
Our EHPE involves two stages, the TIP and Wrist Joints Estimation stage (TW-stage) and the Prior Guided Joints Estimation stage (PG-stage).
In the TW-stage, we employ an Hourglass network \cite{hua2020multipath} associating the features extracted by a pre-trained backbone to predict the positions of TIP and wrist. 
In the PG-stage, we design a dual-branch architecture.
We first perform grid sampling using the TIP and wrist joints predicted by TW-stage, and the feature map produced by the pre-trained backbone to synthesize the graph feature.
Then this graph feature is then fed to Structural Prior based Inference (SPI) Module to inference the remaining joints using the hand structure prior knowledge.
On the other hand, we introduce another branch through a feature enhancement module that independently estimates the full hand joints.
Finally, we sum the outputs from two branches with two learnable weights to get the final output.
Consequently, we adopt a two-stage training strategy to train our EHPE.
We first train TW-stage with only supervision on TIP and wrist joints, while PG-stage in not included.
Then PG-stage is trained with the supervision on full hand joints, while the TW-stage is frozen.
In the inference phase, our EHPE is end-to-end that directly outputs the estimated hand pose with an input image.

Our main contributions are summarized as follow:
\begin{itemize}
    \item We systematically analyze the phenomenon of error accumulation for distal joints in gesture estimation, clarify the importance of the TIP joints under the guidance of structural priors, and design a novel segmented architecture for enhanced hand pose estimation.
    
    \item We design a dual-branch structure in which one branch involves a graphic attention layer of the dynamic topology structure to utilize the hand structure prior with the guidance of TIP and wrist, while the other branch estimates the hand poses from the visual features. 
    
    
    
    \item Extensive experiments demonstrate the effectiveness of the proposed methods. 
    Through comparison on public benchmarks, our EHPE surpasses all compared methods and achieves state-of-the-art performance.
\end{itemize}

\section{Related Work}
Learning hand joint positions from monocular RGB images is a formidable challenge. 
Beyond traditional methods, the majority of current approaches are based on deep learning.
In recent years, researchers have conducted extensive attempts from various perspectives in the field of deep learning.

\subsection{Deep Learning Based Methods}
The research on 3D hand pose estimation has a long history.
Those based on deep learning stood out due to their efficient performance and outstanding results.
Within the field of deep learning, there were numerous methods for hand pose estimation such as depth maps based methods \cite{deng2022recurrent,wan2018dense}, RGB-D based methods \cite{mueller2017real,kazakos2018fusion}, and multi-view images based methods \cite{khaleghi2022multiview}.

As depth information providing more abundant spatial information, there were some deep learning methods used depth map or RGB-D image as input to estimate hand pose.
Tompson \textit{et al.} \cite{tompson2014real} converted depth map into a 3D voxel grid, and joint coordinates were regressed through a 3D CNN. 
However, voxelization introduced a trade-off between resolution and computational cost.
For RGB-D based methods \cite{cai2018weakly}, fusing RGB and RGB-D could combine both texture and geometric cues, but it required the resolution of modality difference issues.
Ge \textit{et al.} \cite{ge2018hand} directly processed the raw point cloud data. 
They utilized an improved PointNet++ to extract local geometric features and combined it with GCN to model the joint relationships. 
This approach enhanced robustness in occluded scenes but was sensitive to noise.

Early research primarily relied on depth sensors, but with the rise of monocular RGB methods \cite{ge2018hand,moon2018v2v,mueller2018ganerated,zheng2019implicit}, researchers have designed more complex network architectures and loss functions \cite{zheng2024quad,zheng2022constrained} to address the issue of lack of depth information.
There were various approaches within methods that took monocular images as input data.
One way was to predict the three-dimensional pose of the hand directly from image data. 
This approach learnt to capture the mapping relationship between hand poses and image features from a large amount of training data.
HandTailor \cite{lv2021handtailor} proposed a network, which included a 3D hand estimation module and an optimization module. 
Different from the existing learning-based monocular RGB-input approaches, Cai \textit{et al.} \cite{cai20203d} introduced a method that utilized synthetic data and weakly labeled RGB images, directly predicting hand poses through a regression network.
Recent studies \cite{jiang2024evhandpose,tu2023consistent,pavlakos2024reconstructing,zhou2024simple} focused on the use of deep learning technology to optimize the accuracy of the estimation of three-dimensional hand poses.
The other way focused on identifying hand keypoints or regions of interest within an image, and then estimated the 3D pose by localizing these keypoints. 
Based on multi-stage joint prediction, Tome \textit{et al.} \cite{tome2017lifting} utilized a heatmap-based representation for the 2D detection of joints, followed by a network to predict the 3D coordinates. 
Keypoint Transformer \cite{hampali2022keypoint} detected keypoints, which were potential joint positions in the image, by predicting a single heatmap.
Then it predicted the 3D hand poses using a cross-attention module, which selected keypoints associated with each of the hand joints.
HandTrackNet \cite{chen2023tracking} proposed a net which was a hand joint tracking network based on point clouds for the first time. 
However, these aforementioned methods \cite{goodfellow2020generative,haiderbhai2020pix2xray} did not escape the constraints imposed by the hand model's topological structure.

Although the methods mentioned above could still learn the relationships between joints through training of deep learning models, they typically did not make use of prior geometric and kinematic information about the hand, which might lead to inaccuracies in estimating complex hand poses or occlusion situations.
Although some methods have employed GCN, they still utilized a fixed adjacency matrix that failed to accommodate the varying hand image environments.

\subsection{Prior Knowledge of Hand Structure}
Parameterized hand models (such as MANO) \cite{MANO:SIGGRAPHASIA:2017} encoded hand pose and shape using low-dimensional parameters, providing prior anatomical knowledge for pose estimation.
Spurr \textit{et al.} \cite{spurr2020weakly} proposed a set of new losses that constrained the predictions of the neural network within the range of 3D hand configurations that were biomechanically feasible. 
He \textit{et al.} \cite{he20213d} proposed using the output of a parameterized hand model as an initial pose prior.
This method incorporated prior knowledge into the generated hand pose distribution through an adversarial learning framework, significantly enhancing the physiological plausibility of the pose under monocular RGB input.
The proposal of hand parameterized models also suggested that hand topology could play a important role in hand pose estimation.
This inspired the widespread application of graph neural networks.
Ren \textit{et al.} \cite{ren2023decoupled} proposed a modeling approach for the spatial relationship between two hands using compact and semantically explicited joint nodes, which could leverage prior knowledge of the hand bone structure.

While existing methods have recognized the importance of hand structural priors and attempted to model them using GCNs, these approaches primarily treated the hand as a monolithic entity without delving into the internal structural hierarchy to investigate the quantitative contributions of different hand joints to these priors.
On this basis, we estimated the coordinates of the TIP and wrist individually prior to utilizing the structural priors and employed them as known conditions for the priors to better estimate the overall hand pose.

\section{Proposed Method}
\subsection{Overview}

\begin{figure*}[!htp]
  \includegraphics[width=\textwidth]{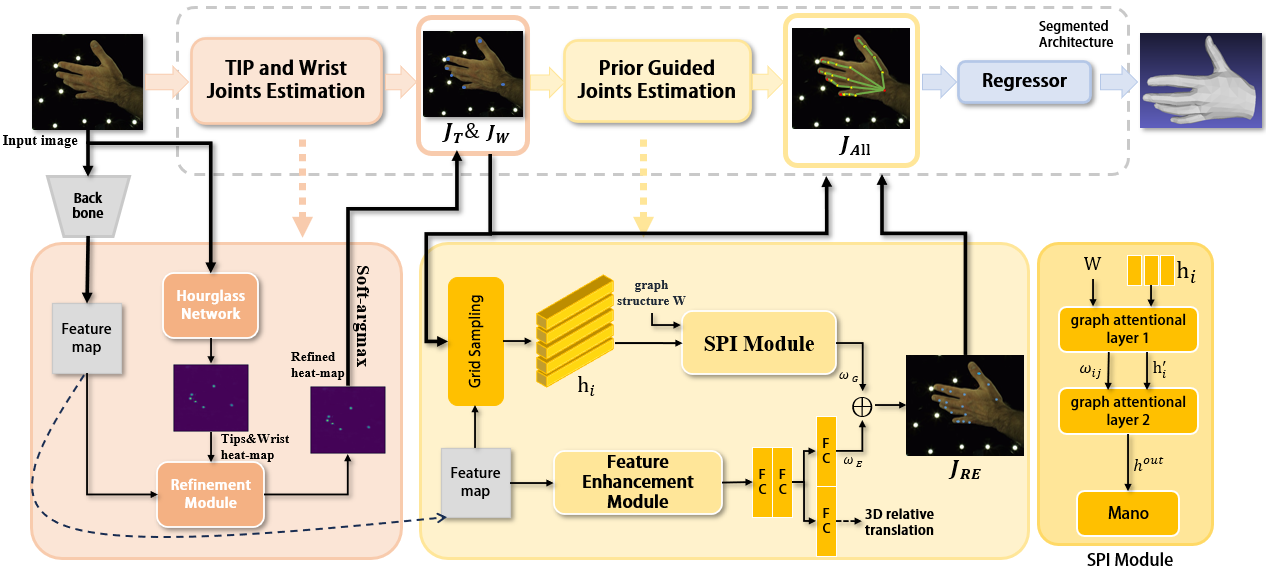}
  \caption{Overview of our segmented architecture framework.
Our framework processes monocular RGB hand image in the TIP and Wrist Joints Estimation stage (TW-stage) and generates joint heatmap under the supervision of heatmap loss, ultimately predicting coordinates of the TIP and wrist joints. 
In the Prior Guided Joints Estimation stage (PG-stage), the network receives the predicted coordinates of the TIP and wrist joints as well as the hand image feature map from the TW-stage. 
PG-stage performs joints inference based on structural priors and joint prediction based on image features simultaneously on the input.
Finally, a weighted fusion of the outputs from the structure yields the final coordinates.}
  \label{icml-over_2}
\end{figure*}

Figure \ref{icml-over_2} provides an overall architecture of our EHPE. 
The proposed EHPE involves a TIP and Wrist Joints Estimation stage (TW-stage) for TIP and wrist joint positions estimating and a Prior Guided Joints Estimation stage (PG-stage) for estimating remaining hand joint positions on the basis of TIP and wrist positions which output from the TW-stage.
\setlength{\intextsep}{5pt}
\setlength{\textfloatsep}{5pt} 

Specifically, an RGB image $\mathbf{I} \in \mathbb{R}^{H \times W \times 3}$ centered on the hand is used as input. 
$H$ = 256 and $W$ = 256 represent height and width of the input image.
A ResNet processes the input image $\mathbf{I}$ to extract feature map $\mathbf{F} \in \mathbb{R}^{h\times w \times C}$,where $h$ = 8, $w$ = 8, $C$ = 2048, which is then fed into a refinement module to generate a refined 2D heatmap. 
To enhance the initial predictions, a refinement module composed of multiple convolutional layers with residual connections combines the 2D heatmap with the image feature maps.
The refined output is reshaped into a 2.5D heatmap with dimensions $\mathbb{R}^{h \times w \times d \times J}$, where $d = 8$ represents the depth resolution, and $J$ is the number of joints. 
The 2.5D joint coordinates for TIP ($J_T \in \mathbb{R}^{5 \times 3}$) and Wrist ($J_W \in \mathbb{R}^{1 \times 3}$) are extracted from the heatmap using a soft-argmax operation, providing the key initial joint predictions required for the next stage. 

In the PG-stage, the input includes the hand TIP joint $J_{T}$, Wrist joint $J_{W}$, and image feature map from the TW-stage. 
Initially, $J_{T}$ and $J_{W}$ are fed into the Structural Prior Based Inference Module (SPI). 
At the same time, the hand image feature map is input into the Feature Enhancement Module \cite{park2023extract}. 
For each joint point, the corresponding image feature map is located within the network based on its 2D image coordinates $(x, y)$. 
The grid sampling is then used to extract the feature vector of each joint point from the image feature map, forming the image feature $\mathbf{F}_i$ associated with the joint. 
Ultimately, the input features of the joint integrate its own geometric features $(\mathbf{x}, \mathbf{y}, \mathbf{z})$ and the visual features $\mathbf{F}_i$ extracted from the image feature map, which are combined to form the feature vector $\mathbf{H}_i = [x_i, y_i, z_i, \mathbf{F}_i]$. 
The graphic attention further utilizes K-independent attention heads to aggregate features through a multi-head attention mechanism, concatenating the outputs of these heads to obtain the final joint point features.
In the end, the outputs of the Structural Prior Based Inference Module and Feature Enhancement Module are weighted fused, which is then predicted to determine the hand joint $J_{RE}$, where $J_{RE}$ denotes the coordinates of remaining joints.

\subsection{TIP and Wrist Joints Estimation}

Heatmap prediction methods excel in local keypoint extraction, particularly in capturing local details and achieving precise localization. 
In this work, we combine 2D heatmaps with hand image features to construct latent features for predicting the coordinates of the TIP and wrist joints.

For each target joint (such as the TIP and wrist), we generate a heatmap of size $h/4 \times w/4 \times J_{T+W}$, where $J_{T+W}$ denotes the total number of TIP and wrist joints and each pixel represents the probability of the position being the target joint.
The true coordinates of the labeled joint $j$ are $(x_j, y_j)$, and a 2D Gaussian distribution is used to generate the true heatmap $H_{\text{true}}^j$ as follows:
\begin{align} H_{\text{true}}^j(i, j) = \frac{1}{2\pi\sigma_x\sigma_y} e^{-\frac{(x_j - x(i))^2}{2\sigma_x^2} - \frac{(y_j - y(j))^2}{2\sigma_y^2}} \end{align} where $H_{\text{true}}^j$ represents the heatmap matrix with dimensions $h \times w$. 
$x_j$ and $y_j$ are the true coordinates of joint $j$, and $x(i)$ and $y(j)$ are the coordinates of the $i$-th column and $j$-th row in the heatmap, respectively. 
$\sigma_x$ and $\sigma_y$ are the standard deviations of the Gaussian distribution, typically set based on joint accuracy and heatmap resolution.

The image is processed through an Hourglass network to generate the true Gaussian heatmap. 
This process provides probabilistic estimates of each joint in the hand image, producing a 2D heatmap with dimensions $h \times w \times J$, where $J$ is the total number of finger and wrist joints.
To integrate the heatmap with the image feature maps, we use a refinement module consisting of eight residual modules and four max-pooling layers. 
The module combines the 2D heatmap with the image feature maps to generate latent feature vectors $F_p$. 
A $1\times 1$ convolutional layer is then used to adjust the channel dimensions, outputting a feature map of size $d \times J$, where $d$ is the discrete depth and $J$ is the number of joints.

The processed features are reshaped into a 2.5D refined heatmap with dimensions $R^{h \times w \times d \times J}$, where $h$ and $w$ are the height and width of the heatmap, $d=8$ represents the discrete depth size, and $J$ indicates the number of joints.
This 2.5D heatmap represents the probability distribution of each joint in the depth direction, enhancing the accuracy of predicting joint coordinates in 3D space.

After obtaining the 2.5D heatmap, the soft-argmax operation is applied to extract the 2.5D coordinates of the TIP and wrist joints from the heatmap. 
For each joint $j$, the 2.5D coordinates $(\hat{x}_j, \hat{y}_j, \hat{z}j)$ are computed as:
\begin{align} (\hat{x}_j, \hat{y}_j, \hat{z}_j) = \sum_{x=1}^{h} \sum_{y=1}^{w} \sum_{d=1}^{8} (x, y, d) \cdot \text{softmax}(H_{\text{pred}}^j(x, y, d)) \end{align} Using the softmax operation, the pixel values in the heatmap are normalized into a probability distribution, enabling a weighted average estimation of each joint's position in the $(x, y, d)$ space. 
This approach helps accurately capture the joint depth information, improving prediction accuracy.

The network is trained jointly with heatmap loss $L_{H}$, Euclidean distance loss $L_{ED}$, and regularization loss $L_{\text{R}}$ during the extraction of joint positions via heatmaps.

\noindent\textbf{Heatmap Loss.} To explicitly constrain the heatmap generated by the model to align with the ground-truth distribution we introduce the MSE loss function.
\begin{align} L_{H} = \frac{1}{J} \sum_{j=1}^{J} \sum_{x=1}^{h} \sum_{y=1}^{w} \left( H_{\text{true}}^j(x, y) - H_{\text{pred}}^j(x, y) \right)^2 \end{align} where $J$ denotes the total number of joint points, and $H_{\text{true}}^j$ and $H_{\text{pred}}^j$ represent the heatmaps of the $j$ th joint point under the true and predicted values, respectively.

\noindent\textbf{Euclidean Distance Loss.} We introduce this loss function to measure the Euclidean distance between predicted and ground-truth joint coordinates, enabling the model to accurately localize each joint in spatial domain.
\begin{align} L_{ED} = \frac{1}{J} \sum_{j=1}^{J} \left( (\hat{x}_j - x_j)^2 + (\hat{y}_j - y_j)^2 + (\hat{z}_j - z_j)^2 \right) \end{align}
where $(\hat{x}_j, \hat{y}_j, \hat{z}_j)$ represents the coordinate of the $j$ th joint predicted by the model, while $(x_j, y_j, z_j)$ denotes the ground truth.

\noindent\textbf{Regularization Loss.} We introduce this loss function to prevent model overfitting and enhance generalization capability.
\begin{align} L_{\text{R}} = \sum_k |W_k|_1 \end{align}
where $W_k$ denotes the $k$ th weight matrix in the model.

\noindent\textbf{Total1 Loss.}The total loss for the TW-stage is computed as: 
\begin{align} L_{\text{total1}} = \lambda_{H} L_{H} + \lambda_{ED} L_{ED} + \lambda_{R} L_{\text{R}} \end{align}
where $\lambda_{H}$, $\lambda_{ED}$ and $\lambda_{R}$ are the weights for the respective loss terms. 
Specifically, we set $\lambda_{H}$ = $3$, $\lambda_{ED}$ = $10^{-2}$ and $\lambda_{R} = 10^{-2} $ in our experiments to balance losses.

\subsection{Prior Guided Joints Estimation}

In this stage, we combine the predicted positions of TIP and wrist joints from the TW-stage with the hand image feature map extracted by the backbone. 
Guided by the predicted joint positions $(\mathbf{x}, \mathbf{y}, \mathbf{z})$, we perform grid sampling on the image feature map. 
For each joint $i$, the feature vector $\mathbf{h}_i = [x_i, y_i, z_i, \mathbf{F}_i]$ is formed by combining the geometric features $(x_i, y_i, z_i)$ and the visual features $\mathbf{F}_i$ extracted from the feature maps. 
This ensures that each joint's feature vector contains both spatial and visual information, providing more comprehensive input to the Structural Prior Based Inference Module, which helps the model better capture the relationships between joints.
\setlength{\intextsep}{3pt}
\setlength{\textfloatsep}{3pt} 
\begin{table*}[htbp]
  \caption{Quantitative comparison with state-of-the-arts methods on FreiHAND dataset. 
The best are shown in bold.}
  \label{table1}
  \begin{center}
  \begin{tabular*}{0.8\linewidth}{@{}@{\extracolsep{\fill}}lllccccc@{}}
    \toprule
        Methods  &Source  & Backbone & PA-MPJPE &PA-MPVPE &F@05 &F@15 &FPS \\
        \midrule
        METRO \cite{zhang2019end} &ICCV'19&HRNet &6.7     &6.8 &0.717&0.981&23         \\
        simpleHand \cite{zhou2024simple} &CVPR'24&HRNet &5.8     &6.1 &0.766&0.986&28              \\
        MobRecon \cite{chen2022mobrecon} &CVPR'22&DenseStack &6.9 &7.2 &0.717  &0.981  & 70     \\
        Tang \textit{et al.} \cite{tang2021towards} &ICCV'21&ResNet50 &6.7  &6.7  &0.724&0.981&39          \\
        IntagHand \cite{li2022interacting} &CVPR'22&ResNet50    &7.5   &7.6&0.692 &0.977 &25          \\
        EANet \cite{park2023extract} &ICCV'23&ResNet50    &5.9     &6.2 &0.771 &0.985 &21        \\
        FastViT \cite{vasu2023fastvit} &ICCV'23&FastViT-MA36 &6.6    &6.7  &0.722 &0.981&75    \\
        HaMeR \cite{pavlakos2024reconstructing} &CVPR'24&ViT &6.0 &\textbf{5.7} &0.785 &0.990 &-  \\
        \midrule
        EHPE(Ours) & &ResNet50 &\textbf{5.7}      &\textbf{5.9} &0.788 &0.986 &28           \\
        EHPE(Ours)  &&HRNet  &5.6  &5.9   &0.787   &0.988    &20   \\
        EHPE(Ours)  &&FastViT-MA36 &\textbf{5.6}      &\textbf{5.7} &0.793 &0.991 &59                           \\
        \bottomrule
\end{tabular*}
\end{center}
\end{table*}

\subsubsection{Feature Enhancement Module} Unlike SPI, the focus of the Feature Enhancement Module lies in processing the visual features of the images, that is, the input feature maps.
Due to the lack of depth information in visual features extracted from monocular RGB images, they are highly ambiguous and confusing.
To address this, we propose a Feature Enhancement Module, primarily composed of self-attention transformers \cite{vaswani2017attention} and cross-attention transformers \cite{vaswani2017attention}.
The dimensions of $\mathbf{q}_{SA}, \mathbf{k}_{SA}, \mathbf{v}_{SA}$ are all $(\mathbf{hw + hw +1} )\times 512$.
\begin{align}
    \text{Attn}(\mathbf{q}_{SA}, \mathbf{k}_{SA}, \mathbf{v}_{SA}) = \text{softmax}\left( \frac{\mathbf{q}_{SA} \mathbf{k}_{SA}^T}{\sqrt{d_{k_{SA}}}} \right) \mathbf{v}_{SA}
    \label{eq:sa}
\end{align}
\subsubsection{ Structural Prior Based Inference Module}
We propose a dynamic graph attention mechanism \cite{velivckovic2017graph} to model the flexible structural relationships between hand joints.
Traditional graph attention networks rely on a fixed adjacency matrix, whose physical meaning lies in encoding hand anatomical constraints through the graph structure.
However, the flexibility of hand joints leads to diverse hand poses.
In different hand poses (such as occlusion or interaction), each edge connecting the joint points plays varying roles in prediction.
In such cases, using a fixed-weight adjacency matrix would fail to adapt to these dynamic variations in hand poses. 

\noindent\textbf{GAT with Dynamic Edge Weights}
Building upon this, we adopt a two-layer graph attention to extract both local and global structural features of the hand. 
The first-layer attention aggregates features from immediate neighbors, capturing local interactions that define finger-level dependencies. 
In contrast, the second-layer attention extends aggregation to two-hop neighbors, modeling long-range dependencies that are critical for understanding overall hand gestures.
For each node $i$, the attention weight $\alpha_{ij}$ between node $i$ and its adjacent node $j$ is computed, followed by feature aggregation:
\begin{align}
h_i^{\text{out}} = \text{concat}_{k=1}^K \sum_{j \in \mathcal{N}_k(i)} \alpha_{ij}^{(k)} W^{(k)} h_j
\end{align}%
where $\mathcal{N}_k(i)$ denotes the neighborhood of node $i$ on the $k$ th scale, K=8. 
This design ensures that both local details and broader spatial relationships are effectively captured.
\begin{align}
\alpha_{ij} = \text{softmax}( \text{LeakyReLU} ( \mathbf{a}^\top [\mathbf{W} \mathbf{h}_i \Vert \mathbf{W} \mathbf{h}_j])) 
\end{align}%
The attention weight $\alpha_{ij}$ is computed using a dot-product attention mechanism. First, attention scores are calculated, and then normalized using softmax to obtain dynamically adjusted edge weights.

To further enhance representation learning, we incorporate cross-layer feature fusion to maintain fine-grained associations across layers:
\begin{align}
h_i^{(l+1)} = \sigma\left(W^{(l+1)} h_i^{(l)} + W_{\text{skip}}^{(l)} h_i^{(0)}\right)
\end{align}%
where $h_i^{(0)}$ represents the initial input features, and $W_{\text{skip}}^{(l)}$ is the learned weight matrix for cross-layer connections. 
This fusion strategy ensures that the model can access the original input features at all layers, effectively preserving the fine-grained association relationships between nodes.

After extracting enhanced node features, we infer joint coordinates using outputs from the Structural Prior Based Inference Module (SPI) and Feature Enhancement Module (FEM). 
The final coordinates for each node $i$ are computed as:
\begin{align}
\hat{c}_i = \omega_{\text{G}} SPI_i^{\text{out}} + \omega_{\text{E}} FEM_i^{\text{out}}
\end{align}%
where $\omega_{\text{G}} \in \mathbb{R}^{21 \times 21}$ and $\omega_{\text{E}} \in \mathbb{R}^{21 \times 1}$ are the learnable weight matrixs.
In the formula, "out" refers to the coordinates obtained after the FC layer.

\noindent\textbf{Position Loss.} To optimize the accuracy of joint point prediction, we minimize the MSE loss between the predicted and ground-truth joint positions:
\begin{align}
L_{\text{P}} = \frac{1}{N} \sum_{i=1}^N \| \hat{c}_i - c_i \|^2
\end{align}%
where $N$ is the total number of nodes, $\hat{c}_i$ and $c_i$ represent the predicted position and the actual position, respectively.

\noindent\textbf{Edge Loss.} Additionally, to prevent extreme variations in edge weights, we impose a regularization term:
\begin{align}
L_{\text{E}} = \sum_{(i, j) \in E} (\alpha_{ij} - 1)^2
\end{align}%
where $\alpha_{ij}$ is the edge weight between node $i$ and node $j$.

\noindent\textbf{Total2 Loss.} The total loss for the PG-stage is computed as:
\begin{align} L_{\text{total2}} = \lambda_{P} L_{P} + \lambda_{E}L_{E} \end{align}
where $\lambda_{P}$ and $\lambda_{E}$ are hyper-parameters to balance the contribution of each loss.
Specifically, we set $\lambda_{P}$ = $2\times10^{-2}$ and $\lambda_{E}$ = $2\times10^{-1}$ in our experiments. 
These settings are designed to ensure that all losses achieve similar values when training is already finished.
\section{Experiments}
\subsection{Implementation Details}
We implement our model using PyTorch, with the Adam optimizer \cite{Kingma_Ba_2014} and a batch size of 32 per GPU (training on two RTX 3090 GPUs).
The backbone are based on ResNet-50 \cite{he2016deep}, HRNet \cite{sun2019deep} and FastVit-MA36 \cite{vasu2023fastvit} with initial weights pre-trained on ImageNet.

For training on the InterHand2.6M dataset \cite{moon2020interhand2}, we use 20 epochs in the TW-stage, starting with an initial learning rate of $1 \times 10^{-3}$.
In the PG-stage, we use 40 epochs with learning rate annealing at the 15th and 20th epochs, starting with an initial learning rate of $1 \times 10^{-4}$. 
For training on the FreiHAND dataset \cite{zimmermann2019freihand}, we train for 50 epochs in the TW-stage and 100 epochs in the PG-stage. 
The learning rate in the TW-stage is set to $5 \times 10^{-3}$. 
In the PG-stage, the learning rate starts with $5 \times 10^{-4}$, which is reduced to $5 \times 10^{-5}$ after 50 epochs.
All images are resized to $256 \times 256$.

\subsection{Dataset and Metrics}
\textbf{Dataset.} 
We evaluate our method on two challenging hand interaction datasets: InterHand2.6M and FreiHAND. InterHand2.6M consists of 2.6 million images of hand interactions, including 1.36 million training images and 849K test images. 
It covers a variety of single-hand and dual-hand poses, with semi-automatic annotations.
FreiHAND is Comprised of over 130,000 RGB images, each annotated with 21 3D hand joint positions, 3D hand meshes, and camera parameters. 
This dataset includes a diverse set of hand poses, and real-world backgrounds, providing variability and robustness.

\noindent\textbf{Evaluation metrics.} 
Following the prior arts \cite{li2023diffhand,chen2022mobrecon,ren2023end}, we evaluate the performance of 3D hand pose estimation using four metrics: Mean Per Joint Position Error (MPJPE) and Mean Per Vertex Position Error (MPVPE) are used for InterHand2.6M. 
Procrustes-aligned Mean Per Joint Position Error (PA-MPJPE), and Procrustes-aligned Mean Per Vertex Position Error (PA-MPVPE) are used for FreiHAND. 
All metrics are in a millimeter scale. 

\subsection{Comparisons with State-of-the-arts Methods}
\subsubsection{Quantitative Comparisons.} Table \ref{table1} and \ref{table2} show that our freamwork achieves the highest performance on FreiHAND, and Interhand2.6M datasets, respectively.
Following the previous works, we measure MPJPEs and PA-MPJPEs after scale alignment on the results.
When testing errors across various datasets, as the corresponding results are not provided in the manuscripts, we retrain and test the model on the various datasets using their officially released codes.
We compare the performance with previous hand joints estimating methods \cite{li2021two,zhou2024simple,zhang2019end,park2023extract,tang2021towards,vasu2023fastvit,chen2022mobrecon,pavlakos2024reconstructing} on various important datasets.

We first compare our method with other mainstream hand estimation methods on the single-hand dataset FreiHAND, as shown in Table~\ref{table1}. 
We use ResNet and FastViT-MA36 \cite{vasu2023fastvit} as the mainstays for non-real-time and real-time methods, respectively. 
As can be seen from Table~\ref{table1}, whether for non-real-time or real-time methods, our method demonstrates an advantage in the results, although slightly slower than FastViT, we achieve an improvement of 1.0mm in the PA-MPJPE.

\begin{table}[ht]
\caption{Quantitative comparison with state-of-the-arts methods on InterHand2.6M dataset.}
\label{table2}

\begin{tabular*}{0.8\linewidth}{@{}@{\extracolsep{\fill}}llcc@{}}
 \toprule
        Methods   &Source   & MPJPE & MPVPE  \\
        \midrule
        Intershape \cite{Zhang_Wang_Deng_Zhang_Tan_Ma_Wang_2021}  &ICCV'21   &13.07          &14.35         \\
        IntagHand \cite{li2022interacting}  &CVPR'22        &8.79           &9.03          \\
        simpleHand \cite{zhou2024simple} &CVPR'24 &8.14           &9.22        \\
        EANet \cite{park2023extract} &ICCV'23 &5.88 &6.04 \\
        \midrule
        EHPE(Ours)          &    &\textbf{5.73}      &\textbf{5.87}           \\
        \bottomrule
\end{tabular*}
\end{table}
As shown in Table~\ref{table2}, we compare our method with other mainstream methods on the two-handed dataset InterHand2.6M. 
Due to testing on the inter-hand dataset, MPJPE and MPVPE are chosen to retain translation and rotation errors, which allows for a more intuitive assessment of the model's prediction accuracy. 
Firstly, our method significantly outperforms all single-handed methods, which is probably due to their inability to handle complex hand occlusions, whereas our method, thanks to the introduction of EABlock, can adapt well to the two-handed scenario. 
Secondly, compared to two-handed methods, our approach also has an advantage, which can be attributed to our two-stage joint point prediction design.

\begin{figure*}[t]
\vskip 0.2in
\begin{center}
\centerline{\includegraphics[width=\textwidth]{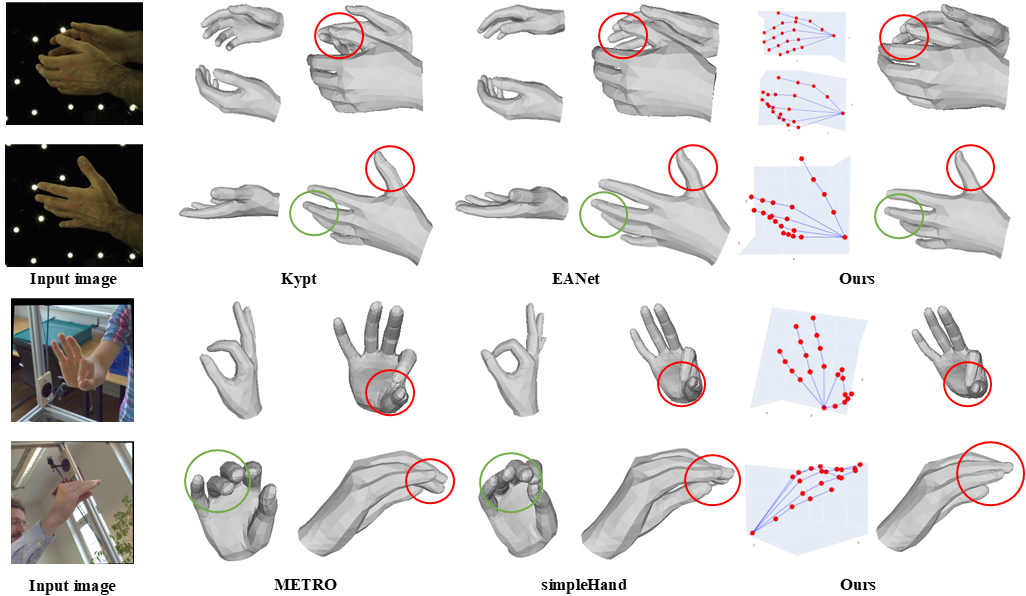}}
\caption{Qualitative comparison between our method and other state-of-the-arts approaches on InterHand2.6M (top) and FreiHAND (bottom).
The red and green circles highlight regions where ours are correct while others are wrong.}
\label{icml-compar}
\end{center}
\vskip -0.2in
\end{figure*}

\subsubsection{Qualitative Results.}
To conduct a qualitative analysis, we compare our method with mainstream approaches on the FreiHAND and InterHand2.6M datasets. 
As illustrated in Figure \ref{icml-compar}, our method demonstrates superior performance on the InterHand2.6M and FreiHAND dataset, generating higher-quality predictions of hand joint points 
compared to previous state-of-the-arts methods, even under complex occlusions and various interaction scenarios.

\subsection{Ablation Studies}
We conduct a series of ablation experiments on the FreiHAND dataset to thoroughly assess the effectiveness of various combinations. 
\setlength{\intextsep}{2pt}
\setlength{\textfloatsep}{2pt}
\begin{figure}[htbp]
\vskip 0.2in
\begin{center}
\centerline{\includegraphics[width=\columnwidth]{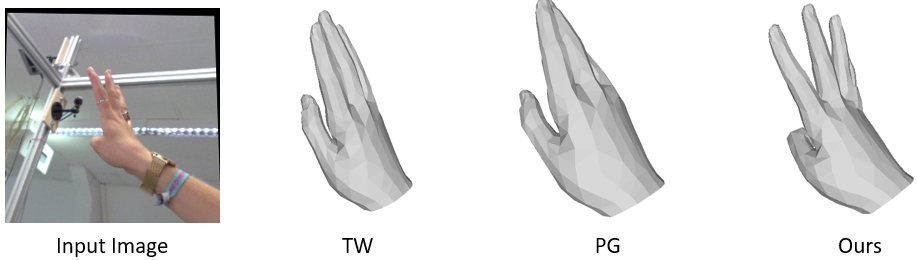}}
\caption{The visual comparisons of various types of segmented architecture.
}
\label{icml-comparE1}
\end{center}
\vskip -0.2in
\end{figure}
For efficiency, all experiments are performed using lightweight model versions. 
In our study, the lightweight version of our proposed model serves as the baseline, and a series of ablation experiments validate the effectiveness of the proposed architecture.

\setlength{\intextsep}{2pt}
\setlength{\textfloatsep}{2pt}
\begin{figure}[htbp]
\begin{center}
\centerline{\includegraphics[width=\columnwidth]{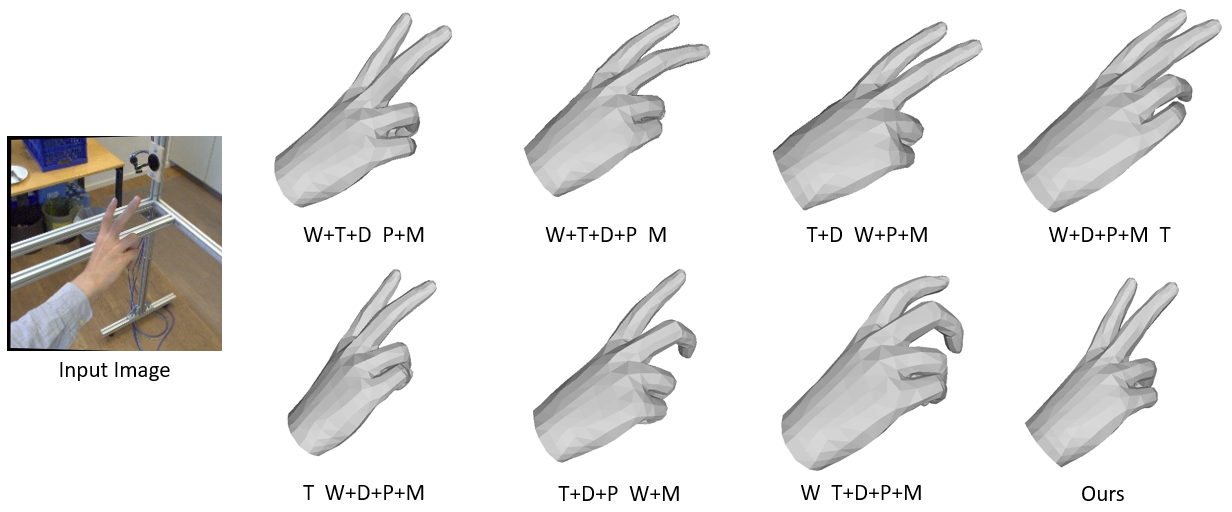}}
\caption{The visual comparisons of various types of segmented architecture.
ere ’T’, ’D’, ’P’, ’M’ and ’W’ in the figure refers to TIP,
DIP, PIP, MCP and Wrist respectively.}
\label{icml-comparE2}
\end{center}
\end{figure}

\subsubsection{Effectiveness of Segmented Architecture.}
In this section, we first investigate the effectiveness of the segment architecture, then demonstrate the rationality of the settings of TW-stage.

\noindent\textbf{Segmented Architecture.}
Table~\ref{table3} and Figure \ref{icml-comparE1} show that the combination of our segmented architecture is optimal. 
In this combination, the TW-stage is designed for TIP and wrist joints estimation, while the PG-stage is employed for estimating the rest joints mainly based on the output of TW-stage. 
Both stages are integral components of our model, and their individual contributions are validated through the ablation experiments.
The top two rows respectively shows estimating full hand joints using TW-stage only and PG-stage only.
Without the hand structure prior provided in PG-stage, the network struggles to handle various hand poses.
Besides, the PG-stage also fails fully utilize the hand structure prior without the guidance from TIP and wrist joints provided by TW-stage.
Notably, it is interesting to see that the TW-stage only and PG-stage only gets almost the same performance, that both of them get a sharp PA-MPJPE/PA-MPVPE reduction of 1.5/1.6 comparing to the full model.
This observation could partially indicate that the guidance of TIP and wrist joints is the keypoint for hand structure prior.
\setlength{\intextsep}{5pt}
\setlength{\textfloatsep}{5pt} 
\begin{table}[tbp]
\caption{Comparison of various combinations of TW-stage and PG-stage on the network.}
\label{table3}
\begin{center}
\begin{tabular*}{0.9\linewidth}{@{}@{\extracolsep{\fill}}ccc@{}}
\toprule
        Stage             & PA-MPJPE & PA-MPVPE  \\
        \midrule
        TW      &7.1 &7.3        \\
        PG     &7.2 &7.3        \\
        \midrule
        EHPE(Ours)                        &\textbf{5.6}          &\textbf{5.7}           \\
        \bottomrule
\end{tabular*}
\end{center}
\end{table}

\noindent \textbf{Settings of TW-stage.}
To demonstrate the rationality of estimating TIP and wrist joints in TW-stage, we conduct experiments of estimating different combinations of hand joints in TW-stage.
As shown in Table \ref{table4}, from the top two rows, estimating additional joints is harmful, because estimating additional joints at one stage clearly sacrifices the accuracy of TIP and wrist joints, thereby leading to an inaccurate hand structure prior.
\begin{table}[htbp]
\caption{Comparison of the results of various allocation combinations of multiple hand joints in the segmented architecture.
Here 'T', 'D', 'P', 'M' and 'W' in the table refers to TIP, DIP, PIP, MCP and Wrist respectively.}
\label{table4}
\begin{center}
\begin{tabular*}{1.0\linewidth}{@{}@{\extracolsep{\fill}}cccc@{}}
\toprule
        TW-stage & PG-stage  & PA-MPJPE & PA-MPVPE  \\
        \midrule
        W+T+D & P+M      &5.8          &5.9           \\
        W+T+D+P & M      &6.2          &6.4           \\   
        T+D & W+P+M       &6.2          &6.3           \\
        T+D+P & W+M      &6.5          &6.7           \\
        T& W+D+P+M                     &5.9 &6.0                 \\
        W & T+D+P+M &7.2  &7.3    \\
        W+D+P+M  & T  &6.4   &6.5    \\
        \midrule
        W+T & D+P+M                       &\textbf{5.6}        &\textbf{5.7}      \\
        \bottomrule
\end{tabular*}
\end{center}
\end{table}
From row 3 to row 5, without wrist joints serving as anchors, the only TIP joints cannot fully stimulate the potential of the hand structure prior.
Moreover, the misleading brought by other joints even leads a negative optimization.
From row 6 to row 7, it's easy to distinguish that TIP is in-replaceable.
The hand structure prior cannot be effectively established without the guidance of TIP joints.
Though other joints could partially contribute to the hand stricture prior, but they cannot achieve similar effectiveness as TIP joints.
\setlength{\intextsep}{2pt}
\setlength{\textfloatsep}{1pt} 
\begin{table}[hbp]
\caption{Comparison of models with various combinations of SPI and FEM.}
\label{table5}
\begin{center}
\begin{sc}
\begin{tabular*}{1.0\linewidth}{@{}@{\extracolsep{\fill}}cccc@{}}
\toprule
        SPI      &FEM      & PA-MPJPE & PA-MPVPE  \\
        \midrule
        $\times$ & $\times$ & 7.1 & 7.3\\
        $\times$  &$\checkmark$      &6.6       &6.6           \\
        $\checkmark$  &$\times$      &6.0       &6.1           \\
        \midrule
        $\checkmark$  &$\checkmark$      &5.6       &5.7           \\
        \bottomrule
\end{tabular*}
\end{sc}
\end{center}
\end{table}
The visualized evidence is provided in Figure \ref{icml-comparE2}.
Through this visual comparison, it's clear that the current settings of TW-stage is the most reasonable.

\subsubsection{Analysis within PG-stage.}
Except for the segmented architecture, the carefully designed PG-stage is another major contribution of this work.
In this part, we conduct a detail analysis for the dual-branch structure and the settings of SPI module.

\noindent \textbf{Dual-branch Structure.}
We first investigate the importance of the dual-branch structure of PG-stage. 
From the quantitative results shown in Table \ref{table5}, both SPI branch and FEM branch could bring considerable performance gains.
This indicates that introducing hand structure prior and enhancing visual features are both helpful for hand pose estimation, while the hand structure prior seems more effective, which is roughly consistent with previous researches \cite{yao2024decoupling, ren2023decoupled}.
Besides, we can also observe that fusing the outputs of two branches achieves further improvement.
This observation indicates that the improvement from visual features and hand structure prior are complementary.

\begin{table}[!htbp]
\caption{Comparison of the results of various graph attentional layers and edge weight setting in SPI.}
\label{table6}
\begin{center}
\begin{tabular*}{1.0\linewidth}{@{}@{\extracolsep{\fill}}cccc@{}}
\toprule
        Edge Weight & Layer Nums            & PA-MPJPE & PA-MPVPE  \\
        \midrule
        fixed & 2 & 6.5 & 6.6 \\
        dynamic & 1      &6.2          &6.3           \\
        dynamic & 3       &5.9          &6.1           \\
        \midrule
        dynamic & 2                      &\textbf{5.6}        &\textbf{5.7}      \\
        \bottomrule
\end{tabular*}
\end{center}
\end{table}
\noindent \textbf{SPI Module.}
The core of SPI module is the use of two graphic attention layers of dynamic edge weights to model the hand structure prior. 
We first investigate the effectiveness of dynamic edge weights applying on graphic attention layers.
As shown in Table \ref{table6}, constructing graphic attention layers with fixed edge weights brings 0.9/0.9 PA-MPJPE/PA-MPVPE reduction.
Compared to the model without SPI module (shown in Table \ref{table5}), the graphic attention layers with fixed edge weights could partially model the hand structure prior.
However, fixed edge weights make it difficult to adapt to various situations and certainly limit the power of graphic attention layers.
In contrast, dynamic edge weights can adaptively adjust the edge weights of invisible joints when dealing with situations like self-occlusion, thereby strengthening the feature transmission of visible joints. 
On the other hand, we also study the effectiveness of using different numbers of graphic attention layers to model the hand structure prior.

As shown in Table \ref{table6}, changing the number of graphic attention layers would certainly affect the model's performance.
The shallow depth (1 layer) cannot well model the hand structure prior, while too many layers may lead to excessive learning of joint relationships, rendering the network incapable of distinguishing between similar joints and result in over-fitting. 
\section{Conclusion}
This paper presents a segmented architecture for the prediction of hand pose. 
During the training process, the framework divides hand joints into TIP and wrist joints as well as other joints. 
In the TW-stage, heatmaps are utilized to predict TIP and wrist joints; in the PG-stage, geometric and image feature are integrated to be the known conditions to forecast remaining joints. 
This segmented architecture effectively distinguishes the contribution of different joints to the hand pose estimation task, reduces the negative impact of distal error accumulation on coordinates prediction, and further explores the role of hand structure prior in the estimation task.
The experiments conducted on two datasets indicate that our method outperforms the previous state-of-the-arts methods.

\begin{acks}
This work is supported by the Key R \& D Program of Zhejiang under Grant No. 2023C01044.
\end{acks}

\newpage
\bibliographystyle{ACM-Reference-Format}
\bibliography{sample-sigconf}










\end{document}